\documentclass[runningheads]{llncs}

\usepackage[T1]{fontenc}
\usepackage{graphicx}
\usepackage{cite}
\usepackage{hyperref} 
\usepackage{color}

\usepackage{booktabs}
\begin{document}

\title{Was Tournament Selection All We Ever Needed? \\ A Critical Reflection on Lexicase Selection}

\titlerunning{Was Tournament Selection All We Ever Needed?}

\author{Alina Geiger \and
Martin Briesch \and
Dominik Sobania \and
Franz Rothlauf} 

\authorrunning{A. Geiger et al.}

\institute{Johannes Gutenberg University, Mainz, Germany\\
\email{\{geiger,briesch,dsobania,rothlauf\}@uni-mainz.de}}

\maketitle            

\begin{abstract}
The success of lexicase selection has led to various extensions, including its combination with down-sampling, which further increased performance. However, recent work found that down-sampling also leads to significant improvements in the performance of tournament selection. This raises the question of whether tournament selection combined with down-sampling is the better choice, given its faster running times. To address this question, we run a set of experiments comparing $\epsilon$-lexicase and tournament selection with different down-sampling techniques on synthetic problems of varying noise levels and problem sizes as well as real-world symbolic regression problems. Overall, we find that down-sampling improves generalization and performance even when compared over the same number of generations. This means that down-sampling is beneficial even with way fewer fitness evaluations. Additionally, down-sampling successfully reduces code growth. We observe that population diversity increases for tournament selection when combined with down-sampling. Further, we find that tournament selection and $\epsilon$-lexicase selection with down-sampling perform similar, while tournament selection is significantly faster. We conclude that tournament selection should be further analyzed and improved in future work instead of only focusing on the improvement of lexicase variants.

\end{abstract}

\section{Introduction}
\label{sec:introduction}

Lexicase selection \cite{Spector.2012, Helmuth.2014} was introduced just over a decade ago and has been quickly adapted in many domains in the field of genetic programming (GP). In comparison to traditional selection methods such as tournament selection, lexicase selection considers each training case individually and not in a compressed form (e.g.~as fitness value). In program synthesis \cite{Helmuth.07082020, Helmuth.2015, Sobania.2023}, the standard version of lexicase was already able to achieve significantly better results than, e.g., tournament selection or fitness-proportionate selection. With extensions such as $\epsilon$-lexicase selection \cite{LaCava.2016, LaCava.2019}, this success could also be extended to symbolic regression \cite{Orzechowski.07022018}, where standard lexicase cannot be used effectively. 

A few years ago, a combination of lexicase and random down-sampling strategies ~\cite{geiger.2023, Hernandez.2019, Ferguson.2020} further improved performance. Instead of considering all available training cases in each generation of a GP run, down-sampling based selection methods use only a random subset, which significantly speeds up the execution time per generation \cite{Hernandez.2019}. Boldi et al. \cite{boldi2024informed} improved this na\"ive random approach by introducing  informed down-sampled lexicase selection which leverages population statistics to select subsets of relevant training cases. 

However, recent studies \cite{geiger2024lexicase, Goncalves.2012} show that a combination of down-sampling with tournament selection reduces overfitting and bloat and also leads to high quality results. Considering this previous work, the question arises: was tournament selection all we ever needed? 

This paper addresses this question in detail and compares $\epsilon$-lexicase selection and tournament selection with both random and informed down-sampling. We analyze the performance, generalization behavior, diversity as well as size of the generated solutions on several synthetic and real-world symbolic regression problems. For the synthetic problems, we analyze different noise levels and different problem sizes to understand if performance is related to those problem characteristics. We compare all methods over the same number of generations instead of a given evaluation budget, because we assume that down-sampling is beneficial even with a much lower number of fitness evaluations. 

We find that down-sampling improves the performance of both selection methods even when using only a small fraction of the fitness evaluations. The larger the problem is, the stronger the performance improvement. The increase in performance comes along with  better generalization behavior when using down-sampling techniques. Moreover, we find that the improvements using down-sampling techniques are stronger for tournament selection than for $\epsilon$-lexicase selection. Tournament selection combined with down-sampling performs similar to $\epsilon$-lexicase selection, with tournament selection being considerably faster. 
With respect to population dynamics, down-sampling reduces code growth, especially when combined with tournament selection. Further, we find that down-sampling increases population diversity when combined with tournament selection. We conclude that down-sampling techniques close the gap between tournament selection and $\epsilon$-lexicase selection for symbolic regression problems, making tournament selection the better choice.  

We discuss related work in Sect.~\ref{sec:related_work} and describe our experimental setup including benchmark problems and parameter settings in Sect.~\ref{sec:experimental_setup}. Section~\ref{sec:results} presents our results, followed by a discussion in Sect.~\ref{sec:discussion}. We conclude our work in Sect.~\ref{sec:conclusion}.

\section{Related work}
\label{sec:related_work}

Tournament selection is a commonly used selection method, where $n$ randomly chosen individuals compete in a tournament and the best individual is selected as a parent for the next generation \cite{Poli.2008}. One disadvantage of tournament selection is that the quality of individuals is measured in terms of an aggregated value, leading to a loss of information about the data structure~\cite{Krawiec.2014}. In contrast, lexicase selection~\cite{Spector.2012, Helmuth.2014} evaluates individuals on each training case separately. This is beneficial during search, as individuals that solve certain training cases particularly well are preserved~\cite{Helmuth.2019, Helmuth.2020b, Pantridge.2018}. Additionally, it has been found that lexicase selection is able to maintain a high population diversity in comparison to tournament selection~\cite{Helmuth.2016}. Lexicase selection has been successfully applied in many problem domains~\cite{Helmuth.2014, Helmuth.2015, Sobania.2023, sobania2021generalizabilitymeasure, Aenugu.2019, Moore.2017, Moore.2018, ding2021optimizing}. 

For symbolic regression problems with continuous-valued errors, $\epsilon$-lexicase selection~\cite{LaCava.2016, LaCava.2019} significantly improved the solution quality compared to tournament selection and standard lexicase selection. $\epsilon$-lexicase selection also considers individuals for selection that are near-elite on training cases by applying an $\epsilon$-threshold to standard lexicase selection.

Recently, the combination of lexicase selection with down-sampling improved the solution quality even more in the domain of program synthesis~\cite{Hernandez.2019, Ferguson.2020} and symbolic regression~\cite{geiger.2023,geiger2024comprehensive,geiger2024lexicase}. A na\"ive down-sampling strategy is random down-sampling~\cite{Goncalves.2012, gonccalves2011experiments}, where only a random subset of training cases is used in each generation to evaluate the quality of the individuals.
It has been found that random down-sampling in combination with tournament selection improves performance, reduces overfitting and controls bloat in the domain of symbolic regression~\cite{Goncalves.2012, martinez2017comparison}. However, reduced overfitting was not (or only insignificantly) observed when combining random down-sampling with lexicase selection for program synthesis problems~\cite{Helmuth.2021, Schweim.2022}.

One drawback of random down-sampling is that the random selection of a subset of training cases may lead to the exclusion of important training cases for several generations. Therefore, Boldi et al.~\cite{boldi2024informed} proposed informed down-sampling, which creates more diverse subsets of training cases by using population statistics. Due to the success of informed down-sampling in program synthesis~\cite{boldi2024informed} and symbolic regression~\cite{geiger2024lexicase} its influence has been further analyzed for program synthesis problems~\cite{boldi2024untangling, Boldi2023.static, boldi2023problem}.

Both random and informed down-sampling have been found to improve performance not only for lexicase selection but also (among others) for tournament selection \cite{geiger2024lexicase, boldi2024untangling}.
Therefore, it is unclear if we need lexicase selection or if tournament selection combined with down-sampling is sufficient considering that tournament selection is significantly faster than lexicase selection.

\section{Experimental Setup}
\label{sec:experimental_setup}

In this section, we present the benchmark problems and GP  parameter settings in detail.

\subsection{Benchmark Problems}
\label{subsec:benchmark_problems}

Our experiments are twofold: First, we study synthetic problems where we are able to modify the problem characteristics such as the level of noise and the number of features in order to better understand the effects of down-sampling. Second, we test our conclusions drawn on the synthetic problems on a representative set of real-world benchmark problems. The problem characteristics are described in Table~\ref{tab:problems}.

\begin{table}[h]
    \centering
    \caption{Benchmark problems with the number of features and the number of instances.}
    \label{tab:problems}
    \begin{tabular}{l|c|c|c}
    \toprule
Problem & \# Features & \# Instances & Type\\
\midrule
friedman1            & 10   	            & 100 & synthetic\\
friedman2            & 4                    & 100 & synthetic\\
friedman3            & 4                    & 100 & synthetic\\
210\_cloud	         & 5                    & 108 & real-world\\
230\_machine\_cpu    & 6                    & 209 & real-world\\
207\_autoPrice       & 15                   & 159 & real-world\\
505\_tecator         & 124                  & 240 & real-world\\

\bottomrule
    \end{tabular}
\end{table}

For our experiments with synthetic problems, we generate Friedman datasets as described in~\cite{breiman1996bagging, friedman1991multivariate}. For all Friedman datasets, we study noise levels of 0.0 (no noise), 0.05, 0.1, and 0.15 as in \cite{de2024srbench++}.\footnote{Noise is only applied to the instances used for training. The test set is noise-free.}
Further, we explore the influence of varying numbers of features. Therefore, we generate the \textsc{Friedman1} problem with $5$, $10$, $25$, and $50$ features. It is important to note that only $5$ features are actually correlated with the target output $y$. 

In addition to the synthetic problems, we select 4 real-world regression datasets from PMLB \cite{Olson2017PMLB} with less than 250 instances from diverse domains and with varying numbers of features. We focus on small datasets to finish our experiments in a reasonable amount of time, as we run GP over a large number of generations. Further, this allows us to better observe the generalization gap.

\subsection{Parameter Settings}
\label{subsec:parameter_settings}

Our experiments are implemented using the DEAP Framework \cite{Fortin.2012}. The parameter settings of the tree-based GP runs are shown in Table~\ref{tab:parameter_setting}.

\begin{table}[h]
  \centering
\caption{Parameter settings of our GP approach.}
\label{tab:parameter_setting}
\begin{tabular}{l|r}
 \toprule 
\textbf{Parameter} & \textbf{Value} \\
\midrule
Population size & $500$ \\
Generation limit & $2,000$ \\
Primitive set & $\{\textrm{\textbf{x}}, \textrm{ERC}, +, -, *, \textrm{AQ}, \textrm{sin}, \textrm{cos}, \textrm{neg}\}$ \\
ERC values & $\{-1,0,1\}$ \\
Initialization method & Ramped half-and-half \\
Maximum tree depth & $17$ \\
Crossover probability & $95\%$ \\
Mutation probability & $5\%$ \\
Runs & 30 \\
\bottomrule
\end{tabular}
\end{table}

All experiments are performed using a population size of $N=500$ and a generation limit of $G=2,000$. We perform the evolutionary runs over a larger number of generations to better observe the generalization gap. Our setup differs from previous studies as evolutionary runs including down-sampling are usually given much more generations compared to runs without down-sampling~\cite{geiger.2023, Hernandez.2019}. However, it has been found that down-sampling has positive effects beyond saving evaluation costs per generation such as overfitting and bloat control~\cite{Goncalves.2012}. Therefore, we assume that down-sampling can be beneficial even when compared over a given number of generations. This means that the evolutionary runs with down-sampling use only a fraction of the fitness evaluations used by the ones without down-sampling. 

For each problem, we split the number of instances into $70\%$ training cases and $30\%$ test cases. The down-sampling parameter is set to $d=0.1$~\cite{geiger.2023}, meaning that only $10\%$ of the training cases are used to evaluate the quality of the individuals in each generation. For informed down-sampling, we set the parent sampling rate to $s=0.01$ and the distance calculation scheduling parameter to $k=10$~\cite{boldi2024informed}.

The fitness measure used to compare individuals in the selection process depends on the selection method. Tournament selection compares individuals based on an aggregated value. Therefore, we measure fitness for runs with tournament selection in terms of the mean squared error (MSE). We set the tournament size to $n=7$. $\epsilon$-lexicase selection selects individuals based on their performance on individual training cases. Hence, for runs with $\epsilon$-lexicase selection, the squared errors of the individuals on the training cases are considered during selection. 

In each generation, we save the individual with the lowest MSE on all training cases as the current best solution. This individual is then evaluated on the unseen test cases with respect to the MSE. Based on that, we calculate the generalization gap as the difference between the MSE on the test cases and the MSE on the training cases. Besides performance, we are interested in the influence of down-sampling on population dynamics. Therefore, we measure in each generation the diversity in the population in terms of error diversity, which is defined as the percentage of distinct error vectors in the population. Moreover, we track the median tree size in the population, where tree size is defined as the number of nodes per tree.

\section{Results}
\label{sec:results}

In this section, we present our experimental results. First, we analyze performance, generalization, diversity, and size of the different selection approaches for synthetic regression problems. Second, we extend our findings to real-world problems. A statistical analysis of our results is provided in the supplementary material (see Appendix D) in our Zenodo repository~\cite{zenodo}.

\subsection{Synthetic Problems with Noise}
\label{subsec:results_synthetic}

We first study the performance and population statistics of tournament and $\epsilon$-lexicase selection with different down-sampling techniques on the synthetic Friedman problems with varying degrees of noise. We investigate no down-sampling (\textit{nds}), random down-sampling (\textit{rds}), and informed down-sampling (\textit{ids}) variants of both tournament (\textit{tourn}) and $\epsilon$-lexicase (\textit{$\epsilon$-lex}) selection. All results are shown for 30 runs for each combination of down-sampling and selection method.

\begin{figure}
    \centering
    \includegraphics[width=1.0\linewidth]{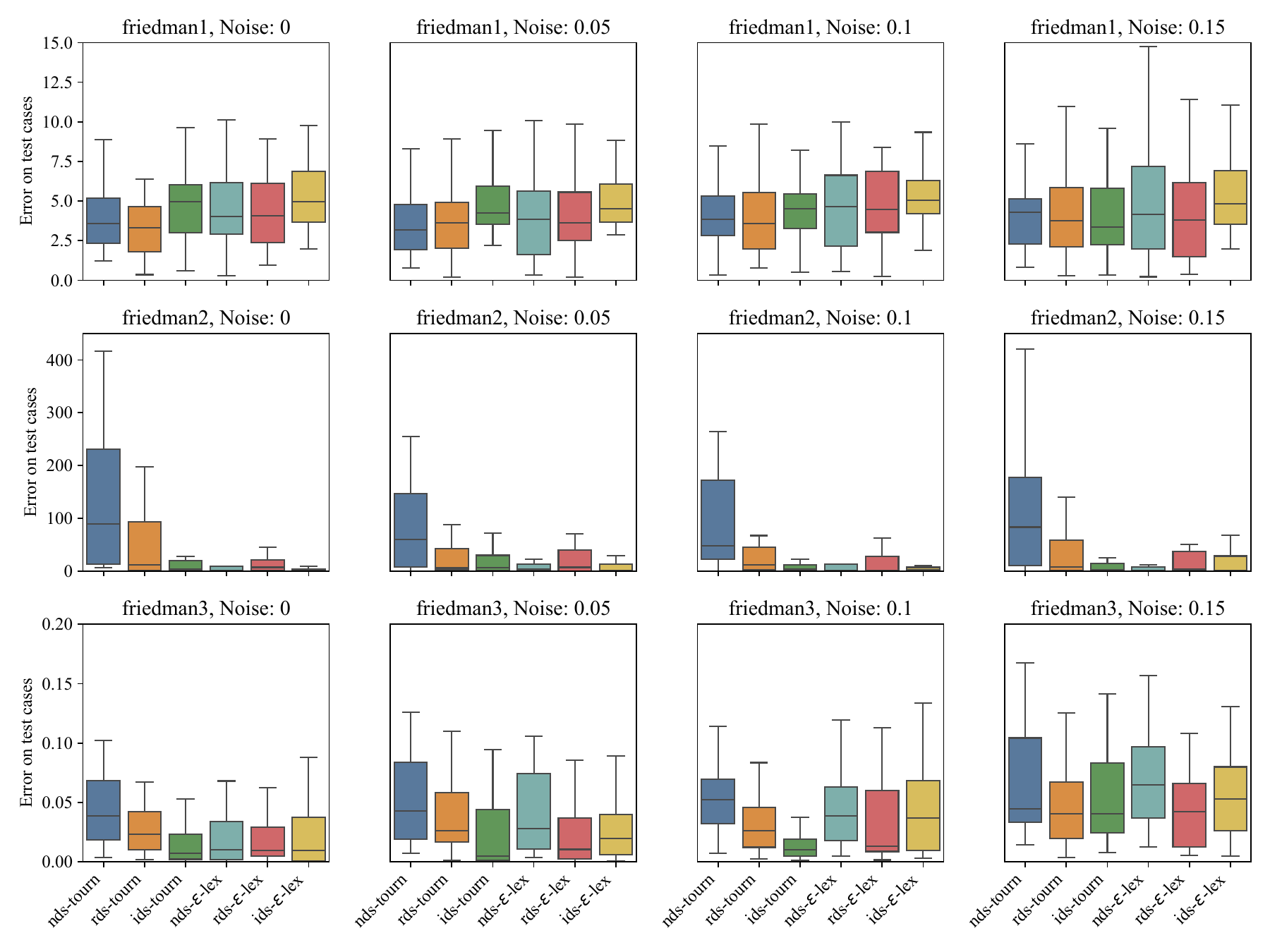}
    \caption{Test fitness of the best individual for the synthetic problems across different noise levels. Outliers are not shown to improve readability.}
    \label{fig:friedman-test_fitness_boxplots}
\end{figure}

Figure~\ref{fig:friedman-test_fitness_boxplots} displays the test error of the best individual found by the respective selection method in an evolutionary run. 

We observe that the down-sampling methods often improve upon or perform equal to the selection method without down-sampling. Only for the \textsc{Friedman1} problem with low noise the down-sampling variants are slightly worse. This is interesting considering that tournament and $\epsilon$-lexicase selection without down-sampling have 10 times more fitness evaluations during training, clearly indicating that down-sampling is beneficial even beyond the saved computational effort. 
Another observation is that tournament selection without down-sampling performs significantly worse than $\epsilon$-lexicase selection without down-sampling on the \textsc{Friedman2} and \textsc{Friedman3} problems. However, when combined with either random or informed down-sampling tournament selection performs equally well as $\epsilon$-lexicase selection at all noise levels. 

\begin{figure}
    \centering
    \includegraphics[width=1.0\linewidth]{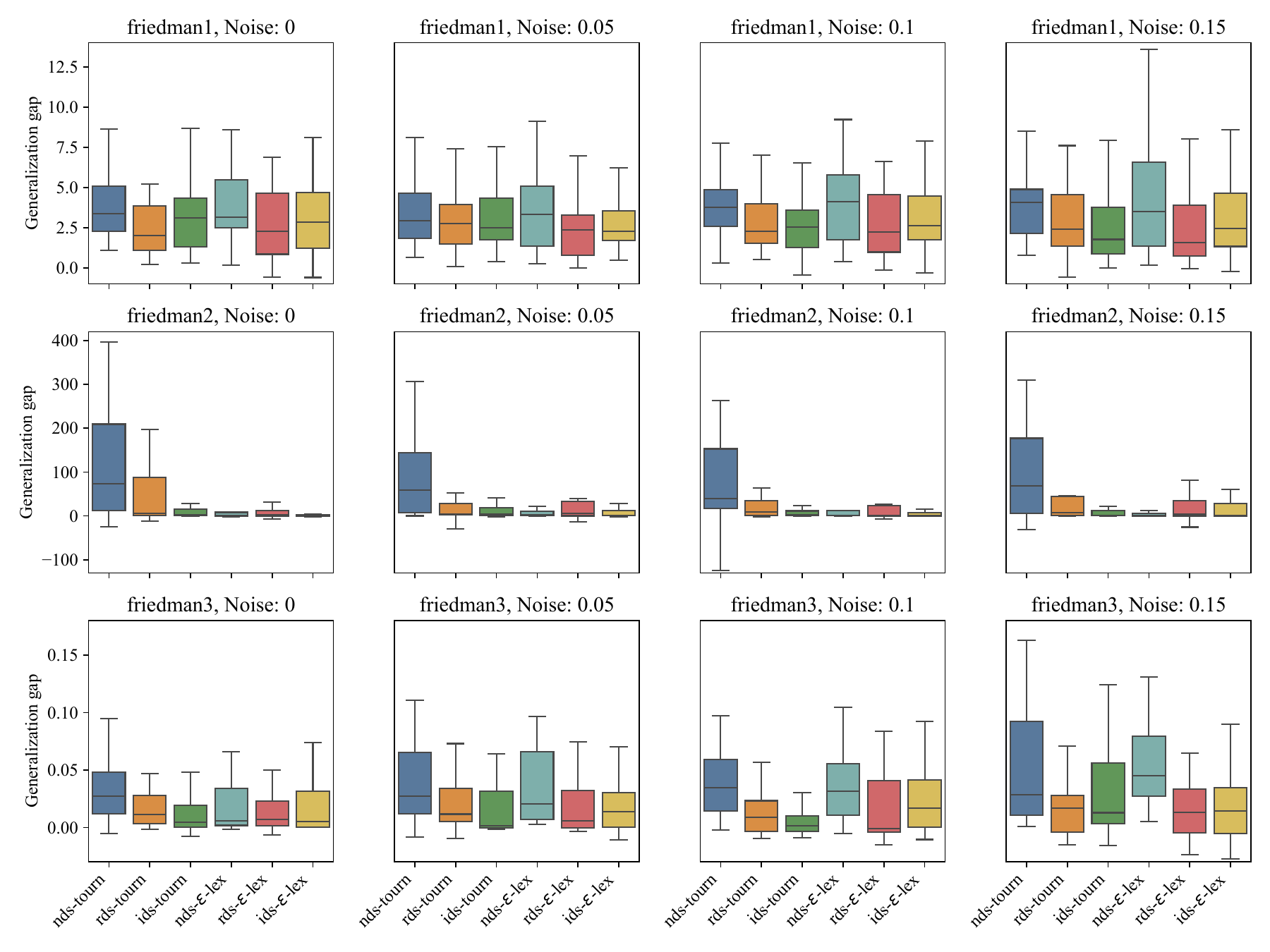}
    \caption{Generalization gap between test and training fitness of the best individual for the synthetic problems across different noise levels. Outliers are not shown to improve readability.}
    \label{fig:friedman-overfitting_boxplots}
\end{figure}

To gain a further understanding of the performance of the different selection methods and down-sampling methods we also investigate their generalization behavior. Figure~\ref{fig:friedman-overfitting_boxplots} displays the generalization gap (difference between test and training error) of the solution per evolutionary run.

In most settings, we observe that down-sampling decreases the generalization gap between training and test error for both selection methods, effectively working as a regularization technique. Especially for tournament selection this effect is apparent. This indicates that both random and informed down-sampling can reduce the generalization gap and negate the difference between both selection methods.

We also provide more details of the training and test error recorded over 2,000 generations in the supplementary material (see Appendix A)~\cite{zenodo}.

\begin{figure}
    \centering
    \includegraphics[width=1.0\linewidth]{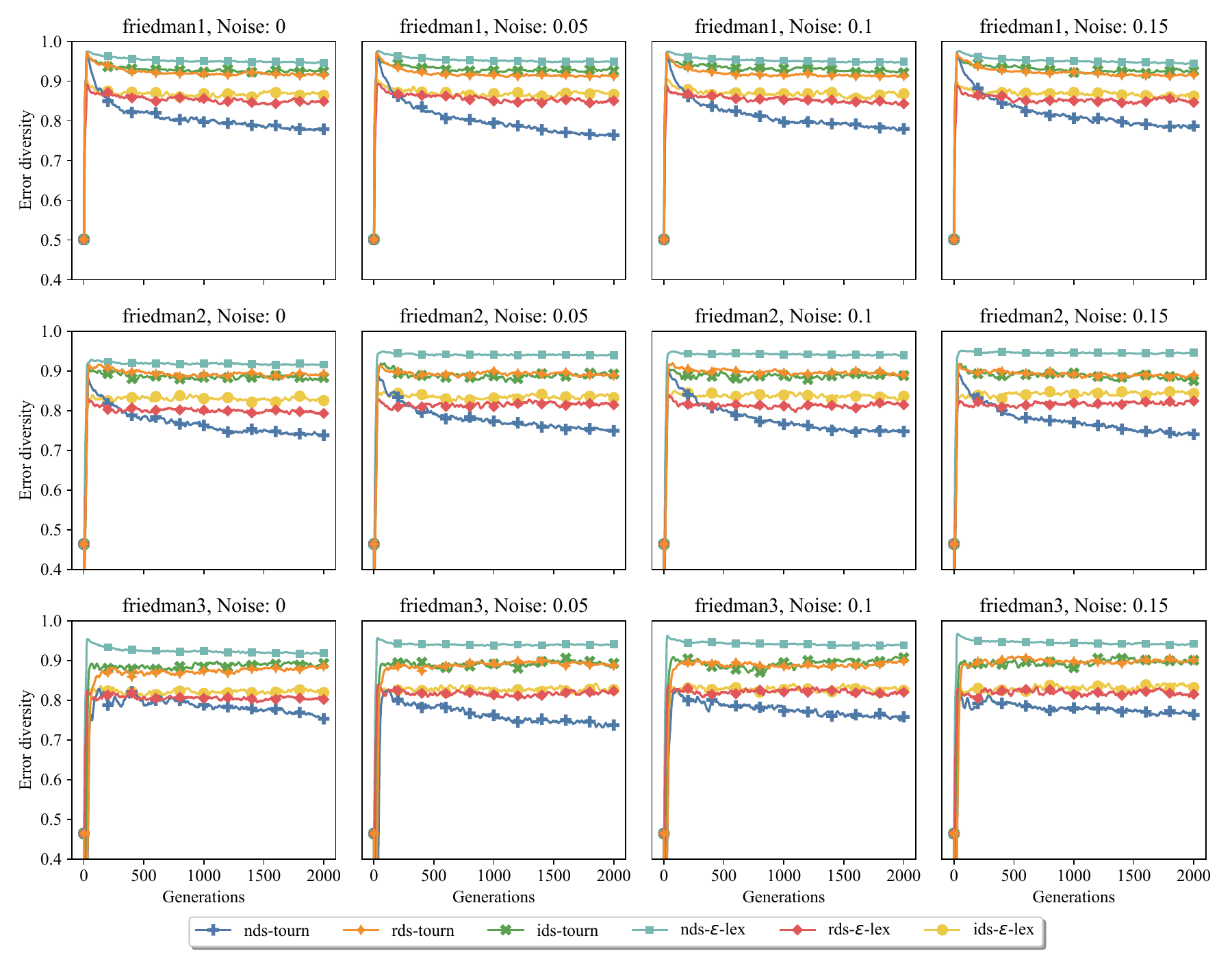}
    \caption{Error diversity of individuals in a population over generations for the synthetic problems across different noise levels. Median over 30 runs is shown.}
    \label{fig:friedman-diversity}
\end{figure}

Another advantage of lexicase selection often mentioned in literature is high diversity preservation within a population \cite{Helmuth.2016}. Figure~\ref{fig:friedman-diversity} displays the diversity (defined as the percentage of distinct error vectors in the population) over generations for the different selection and down-sampling approaches. The graphs are smoothed using a moving average with a window size of 20 generations. Line markers are added every 200 generations for better readability.

As expected, no down-sampling $\epsilon$-lexicase selection exhibits the highest diversity and no down-sampling tournament exhibits the lowest diversity across all problems and noise levels. However, when paired with either random or informed down-sampling the diversity of tournament selection raises significantly, nearly reaching the level of $\epsilon$-lexicase selection and even surpassing random and informed down-sampled $\epsilon$-lexicase selection. This means that down-sampling enables tournament selection to also benefit from higher population diversity.

\begin{figure}
    \centering
    \includegraphics[width=1.0\linewidth]{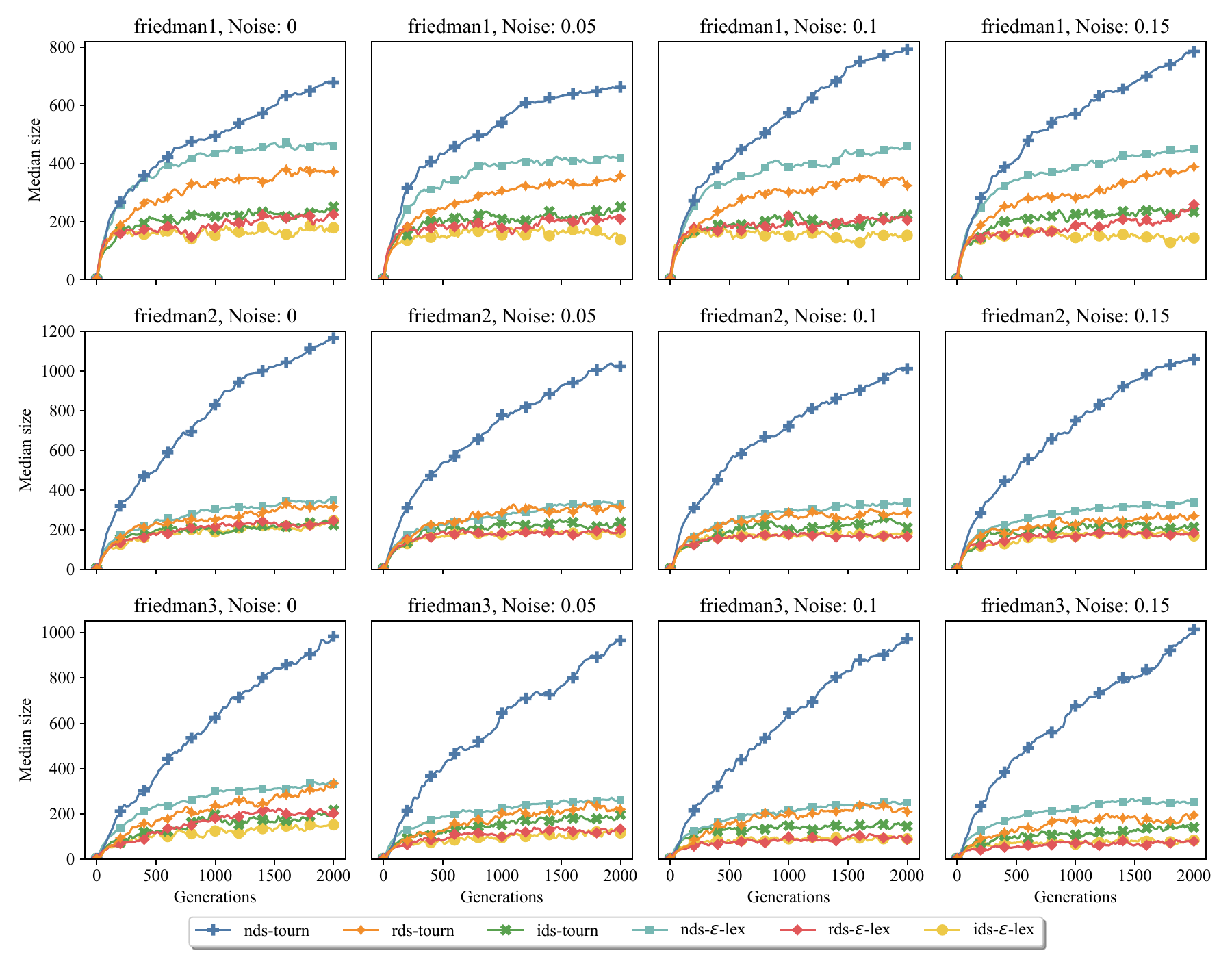}
    \caption{Median size of individuals in a population (measured in tree nodes) over generations for the synthetic problems across different noise levels. Median over 30 runs is shown.}
    \label{fig:friedman-size}
\end{figure}

Lastly, Fig.~\ref{fig:friedman-size} displays the median size of individuals (measured as the number of nodes) within a population. Once again, the graphs are smoothed using a moving average with a window size of 20 generations and line markers are added every 200 generations.
We observe a very strong code growth for tournament selection without down-sampling across all problems and noise levels with a median size of up to 4 times as much tree nodes compared to no down-sampling $\epsilon$-lexicase selection. However, when paired with either random or informed down-sampling this effect is negated similar to the difference in diversity. While the down-sampled $\epsilon$-lexicase selection variants produce the smallest individuals, down-sampled tournament selection produces smaller individuals than no down-sampling $\epsilon$-lexicase selection and comes very close to the down-sampling variants. We also observe that informed down-sampling controls code growth even better than random down-sampling for both tournament and $\epsilon$-lexicase selection. 

Overall we observe that for the synthetic Friedman problems both down-sampling techniques close the gap in performance between tournament and $\epsilon$-lexicase selection. Additionally, down-sampling improves generalization and reduces code growth, especially for tournament selection. Lastly, down-sampling improves diversity for tournament selection and lowers diversity for $\epsilon$-lexicase selection.

\subsection{Synthetic Problems with varying Numbers of Features}
\label{subsec:results_synthetic_features}

We are also interested in how the performance of the selection methods change when we adjust the problem size by increasing or decreasing the feature space. Therefore, we conduct experiments for the \textsc{Friedman1} problem with varying numbers of features (5, 10, 25, 50) and no noise.

\begin{figure}
    \centering
    \includegraphics[width=1\linewidth]{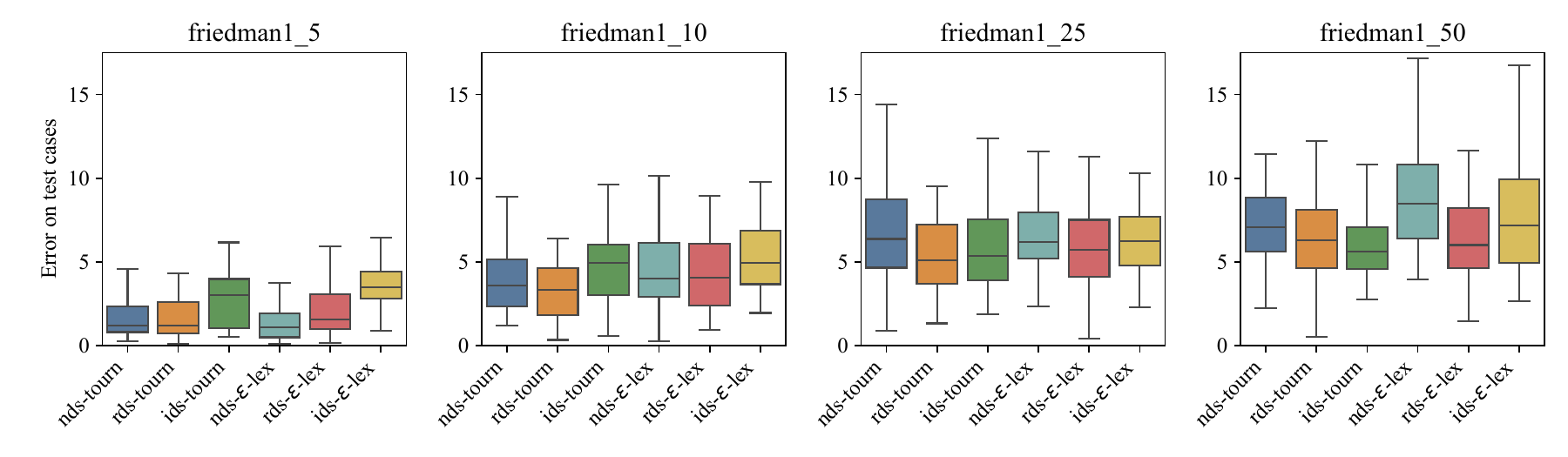}
    \caption{Test fitness of the best individual for the synthetic \textsc{Friedman1} problem with varying numbers of features (5, 10, 25, 50).  Outliers are not shown to improve readability.}
    \label{fig:friedman_features-test_fitness_boxplots}
\end{figure}

Figure~\ref{fig:friedman_features-test_fitness_boxplots} displays the test error of the best individual obtained by the different selection methods. We observe that for the problem instance with only 5 features (that are all correlated to the target variable) both tournament and $\epsilon$-lexicase selection perform better without down-sampling and informed down-sampling performs worst. However, when increasing the problem size, down-sampling performs better than no down-sampling. For the problem instance with 25 features (and only 5 of those are correlated to the target variable), random and informed down-sampled tournament selection achieve the best median error on the test cases.
In the largest problem instance with 50 features, informed down-sampled tournament selection achieves the best median test error. 

\begin{figure}
    \centering
    \includegraphics[width=1\linewidth]{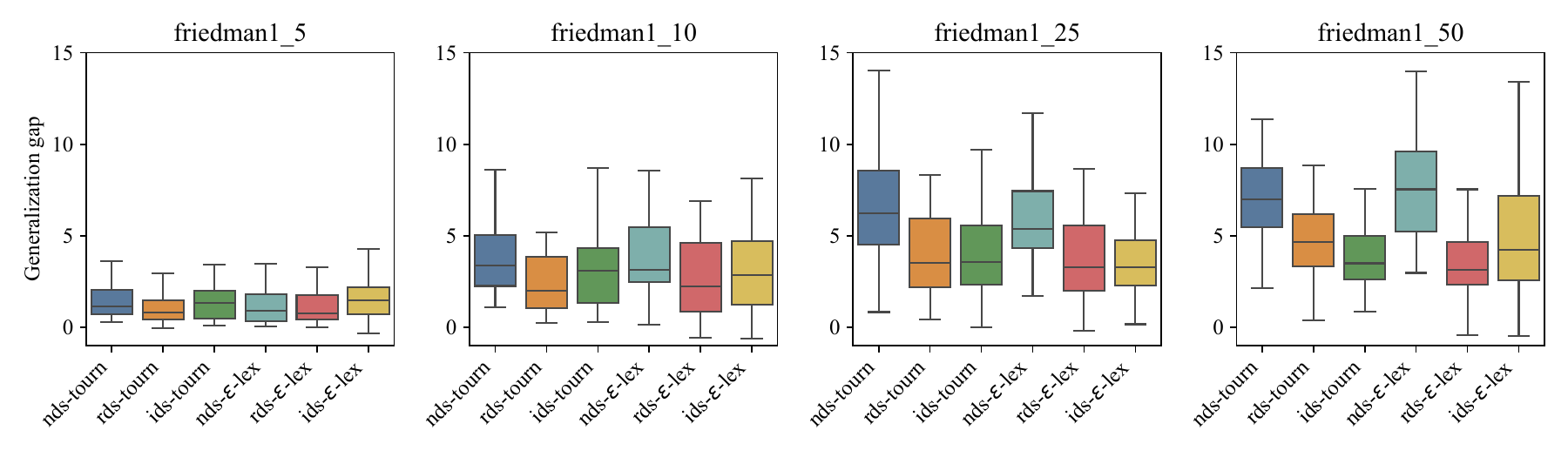}
    \caption{Generalization gap between test and training fitness of the best individual for the synthetic \textsc{Friedman1} problem with varying numbers of features (5, 10, 25, 50).  Outliers are not shown to improve readability.}
    \label{fig:friedman_features-overfitting_boxplots}
\end{figure}

We observe similar findings for the generalization gap displayed in Fig.~\ref{fig:friedman_features-overfitting_boxplots}. There are no major differences in the generalization gaps between tournament and $\epsilon$-lexicase selection as well as no down-sampling and random or informed down-sampling for the small problem instance with 5 variables. However, when increasing the number of features both down-sampling techniques successfully reduce the generalization gap. 

We still observe an increase in diversity and better code growth control for tournament when using down-sampling. However, there is no notable difference between problem sizes (see Appendix C in the supplementary material~\cite{zenodo}).

Overall, these findings indicate that the benefits of down-sampling are more apparent in larger problems and down-sampled tournament selection scales better than $\epsilon$-lexicase selection with increasing problem size.

\subsection{Real-World Problems of Varying Size}
\label{subsec:results_real_world}

\begin{figure}
    \centering
    \includegraphics[width=1\linewidth]{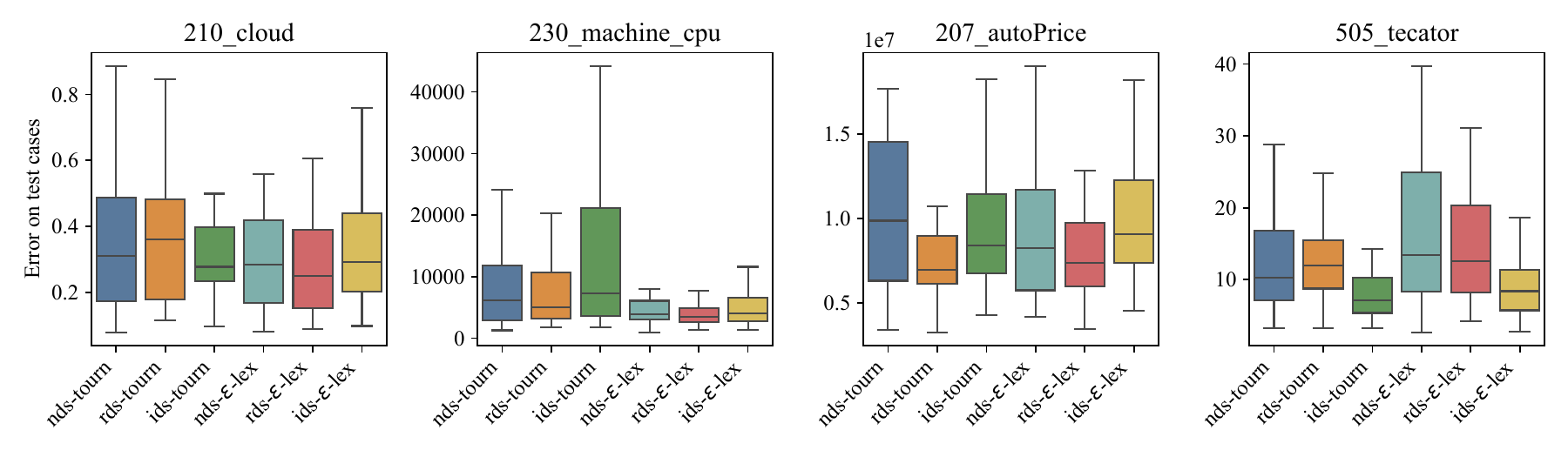}
    \caption{Test fitness of the best individual for the real-world problems.  Outliers are not shown to improve readability.}
    \label{fig:real_world-test_fitness_boxplots}
\end{figure}

Finally, we extend our findings to four real-world problems of varying size.
Figure~\ref{fig:real_world-test_fitness_boxplots} displays the test error of the best individual obtained by each selection method for the four different real-world problems. Both the \textsc{cloud} and \textsc{machine\_cpu} dataset are the smaller problems with 5 and 6 features respectively, followed by \textsc{autoPrice} dataset with 15 features. The \textsc{tecator} dataset is the largest real-world dataset in our experiments with 124 features. 

Similar to our results for the synthetic problems we observe that $\epsilon$-lexicase selection performs slightly better than tournament selection for the two smaller problems (\textsc{cloud} and \textsc{machine\_cpu}) and there is no notable difference between no down-sampling and random or informed down-sampling for those datasets. 
For the larger \textsc{autoprice} problem, down-sampling slightly improves performance for both tournament and $\epsilon$-lexicase selection with random down-sampling tournament selection achieving the best median test error.
For the largest problem (\textsc{tecator}) especially informed down-sampling improves both tournament and $\epsilon$-lexicase when compared to no or random down-sampling and informed down-sampled tournament selection achieves the best median test error.

\begin{figure}
    \centering
    \includegraphics[width=1\linewidth]{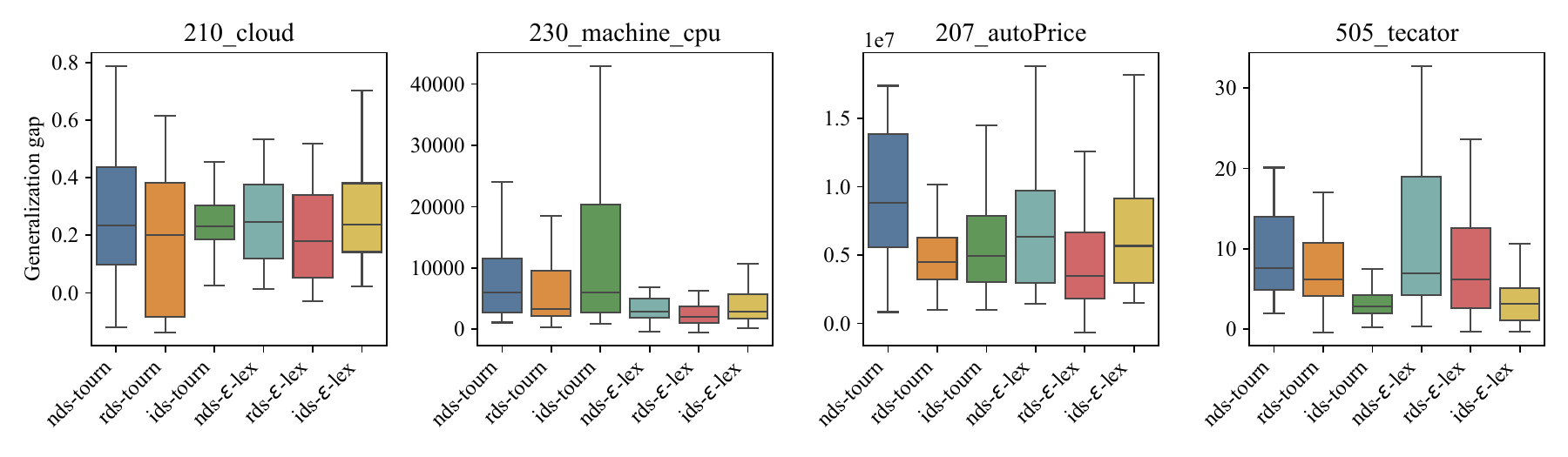}
    \caption{Generalization gap between training and test fitness of the best individual for the real-world problems.  Outliers are not shown to improve readability.}
    \label{fig:real_world-overfitting_boxplots}
\end{figure}

A similar picture is observed for the generalization behavior on those problems as displayed in Fig.~\ref{fig:real_world-overfitting_boxplots}. There are no major differences in the generalization gap between no down-sampling and random or informed down-sampling for the \textsc{cloud} and \textsc{machine\_cpu} dataset and the differences between tournament and $\epsilon$-lexicase selection are only marginal with $\epsilon$-lexicase selection having a slightly lower generalization gap on the \textsc{machine\_cpu} dataset.
However, on the \textsc{autoPrice} dataset, down-sampling clearly reduces the generalization gap, especially for tournament selection. On the \textsc{tecator} dataset, informed down-sampling achieves the smallest generalization gap for both selection methods. Additionally, we observe a higher variance for $\epsilon$-lexicase selection.

More details on the training and test error over generations are provided in the supplementary material (see Appendix B)~\cite{zenodo}, further confirming our results.

\begin{figure}
    \centering
    \includegraphics[width=1\linewidth]{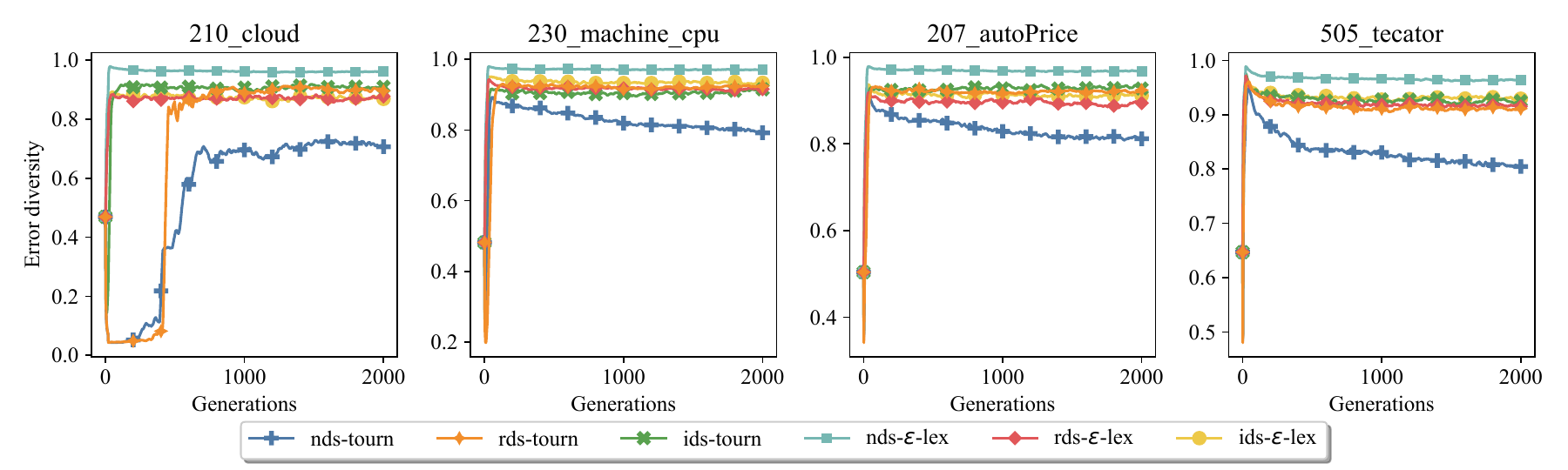}
    \caption{Error diversity of individuals in a population over generations for the real-world problems. Median over 30 runs is shown.}
    \label{fig:real_world-diversity}
\end{figure}

Figure~\ref{fig:real_world-diversity} displays the diversity over generations for the different selection methods. Similar to the synthetic problems we observe that $\epsilon$-lexicase selection exhibits the highest diversity and tournament selection exhibits the lowest diversity. However, when paired with down-sampling the diversity of tournament selection is improved to the level of down-sampled $\epsilon$-lexicase selection regardless of problem size, confirming our results observed for the synthetic problems.

\begin{figure}
    \centering
    \includegraphics[width=1\linewidth]{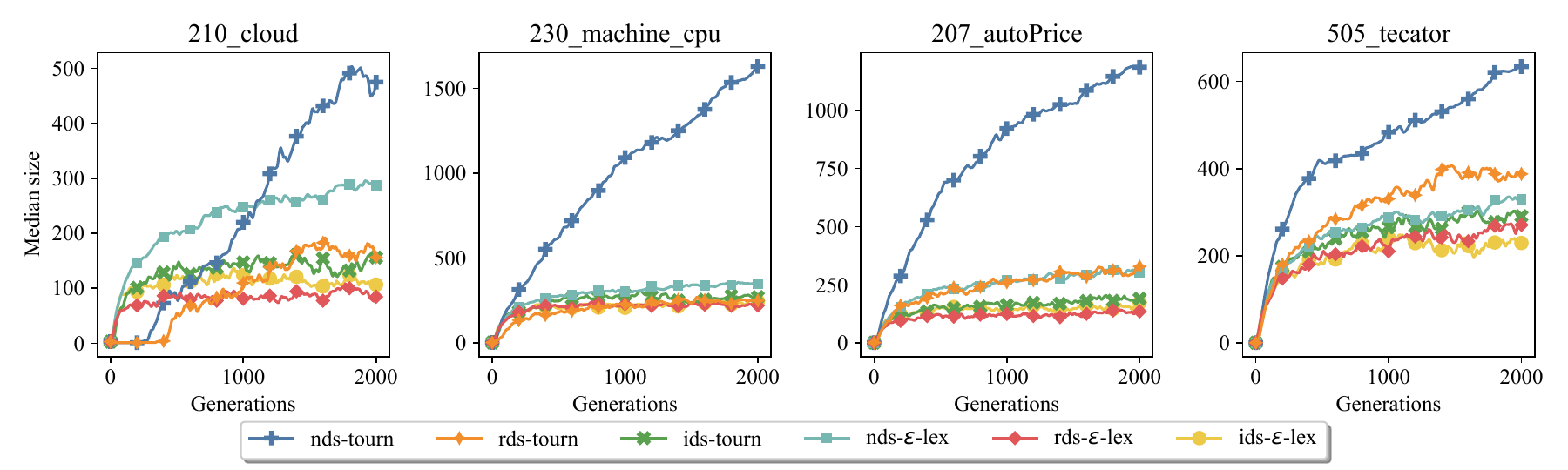}
    \caption{Median size of individuals in a population (measured in tree nodes) over generations for the real-world problems. Median over 30 runs is shown.}
    \label{fig:real_world-size}
\end{figure}

We can also extend our findings regarding code growth control to the real-world datasets. As displayed in Fig.~\ref{fig:real_world-size}, down-sampling successfully reduces code growth for both tournament and $\epsilon$-lexicase selection for all datasets. Especially for tournament selection this is crucial as tournament selection suffers from the most severe code growth in our experiments. With down-sampling (specifically informed down-sampling), however, the code growth of tournament selection mimics that of $\epsilon$-lexicase.

Overall, we can confirm our findings obtained on the synthetic problems also for real-world problems: First, down-sampling is beneficial even when performed for the same number of generations; second, the benefit of down-sampling increases the larger a problem gets and informed down-sampling in particular performs best on larger problems; third, tournament selection with down-sampling scales better to larger problems than $\epsilon$-lexicase; and fourth, down-sampling serves as code growth control and diversity preservation mechanism, providing tournament selection with the major benefits of $\epsilon$-lexicase selection with no added cost.

\section{Discussion}
\label{sec:discussion}

Our results demonstrate that tournament selection and $\epsilon$-lexicase selection perform similar for symbolic regression problems when coupled with random or informed down-sampling techniques. Further, down-sampling also serves as a code growth control mechanism and preserves diversity for evolutionary runs with tournament selection to the degree $\epsilon$-lexicase selection does. However, tournament selection has a way better worst case performance with $\mathcal{O}(TN)$ compared to lexicase selection with $\mathcal{O}(TN^2)$ for selecting $N$ parents using $T$ training cases~\cite{LaCava.2019}. This is also reflected in our empirical data: Tournament selection with down-sampling took around 40\% less time than its $\epsilon$-lexicase counterpart in our experiments.

However, our findings are limited to symbolic regression problems and might not generalize to other problem types. In the domain of program synthesis, other work suggests that lexicase still outperforms tournament selection even when combined with down-sampling \cite{boldi2024untangling}.

Given the results we achieved with the combination of down-sampling and tournament selection as well as work in other domains, it becomes apparent that future research cannot only focus on a single selection method when providing new improvements but should evaluate their improvements holistically.

\section{Conclusion}
\label{sec:conclusion}

Prior work found that lexicase selection is superior to tournament selection in terms of performance and diversity maintenance~\cite{Helmuth.2014,Helmuth.2015, LaCava.2016}. However, down-sampling techniques~\cite{Goncalves.2012,boldi2024informed} improved the performance of both selection methods and it has been found that tournament selection combined with informed down-sampling even surpasses the performance of $\epsilon$-lexicase selection for symbolic regression problems in some cases~\cite{geiger2024lexicase}. With tournament selection being the faster selection method, we raised the question: Was tournament selection all we ever needed?  

We approached this question by comparing $\epsilon$-lexicase and tournament selection with random and informed down-sampling on synthetic as well as real-world symbolic regression problems in terms of performance, generalization behavior, diversity, and solution size. 
We found that down-sampling is even beneficial when comparing methods over the same number of generations. Further, we observed that down-sampling strongly reduces the generalization gap especially for larger problems. The performance improvements using down-sampling techniques are stronger for tournament selection. Additionally, we found that down-sampling with tournament selection and down-sampling with $\epsilon$-lexicase selection perform similar, while down-sampled tournament selection runs significantly faster. Moreover, we found that down-sampling serves as a code growth control, which is especially relevant for tournament selection. Lastly, down-sampling strongly increases the population diversity for the evolutionary runs with tournament selection. 

To answer our question, we argue that tournament selection might be all we ever needed, at least in the domain of symbolic regression, as it performed similar to $\epsilon$-lexicase selection when combined with down-sampling, while being significantly faster than $\epsilon$-lexicase selection. Therefore, we recommend future research to include tournament selection with down-sampling in their studies instead of only focusing on new lexicase variants. Further, tournament selection with down-sampling should be analyzed in other problem domains as well.

\bibliography{main}
\bibliographystyle{splncs04}

\end{document}